\documentclass{bvm} 

\begin{document}

\newcommand{\bvmyear}{2025}

\selectlanguage{english} 

\title{Histologic Dataset of Normal and Atypical Mitotic Figures on Human Breast Cancer (AMi-Br)}


\titlerunning{A Dataset of Normal and Atypical Mitotic Figures}

\author{
	Christof~A. \lname{Bertram} \inst{1}, 
    Viktoria \lname{Weiss} \inst{1},
	Taryn~A. \lname{Donovan} \inst{2}, 
    Sweta \lname{Banerjee} \inst{3},
    Thomas \lname{Conrad} \inst{4},
    Jonas \lname{Ammeling} \inst{6}, 
    Robert \lname{Klopfleisch} \inst{4},
    Christopher \lname{Kaltenecker} \inst{5},
	Marc \lname{Aubreville} \inst{3}
}

\authorrunning{Bertram et al.}

\institute{
\inst{1} University of Veterinary Medicine, Vienna, Austria\\
\inst{2} The Schwarzman Animal Medical Center, New York, USA\\
\inst{3} Flensburg University of Applied Sciences, Flensburg, Germany \\
\inst{4} Freie Universität Berlin, Berlin, Germany  \\
\inst{5} Medical University of Vienna, Vienna, Austria \\
\inst{6} Technische Hochschule Ingolstadt, Ingolstadt, Germany \\
}

\email{christof.bertram@vetmeduni.ac.at}

\maketitle

\begin{abstract}
Assessment of the density of mitotic figures (MFs) in histologic tumor sections is an important prognostic marker for many tumor types, including breast cancer. Recently, it has been reported in multiple works that the quantity of MFs with an atypical morphology (atypical MFs, AMFs) might be an independent prognostic criterion for breast cancer. AMFs are an indicator of mutations in the genes regulating the cell cycle and can lead to aberrant chromosome constitution (aneuploidy) of the tumor cells.
To facilitate further research on this topic using pattern recognition, we present the first ever publicly available dataset of atypical and normal MFs (AMi-Br). For this, we utilized two of the most popular MF datasets (MIDOG 2021 and TUPAC) and subclassified all MFs using a three expert majority vote.
Our final dataset consists of 3,720 MFs, split into 832 AMFs (22.4\%) and 2,888 normal MFs (77.6\%) across all 223 tumor cases in the combined set.
We provide baseline classification experiments to investigate the consistency of the dataset, using a Monte Carlo cross-validation and different strategies to combat class imbalance. We found an averaged balanced accuracy of up to 0.806 when using a patch-level data set split, and up to 0.713 when using a patient-level split.
\end{abstract}

\section{Introduction}

For many human and animal tumors, including human breast cancer, one of the most important tests for assessing patient prognosis is quantification of tumor cell proliferation~\cite{bertram2024mitotic,van2022grading}. 
Dividing cells can be detected in standard histological tumor images (H\&E stain) as mitotic figures (MFs) and are counted within a specific tumor area (mitotic count). Due to relevant inter-rater variability of pathologists in assessment of the mitotic count, deep learning models have been extensively investigated as a potential solution for assistance of the prognostication process~\cite{aubreville2024domain,veta2019predicting,aubreville2023mitosis}. For this purpose of model development, several challenge datasets have been created and made publicly available, most importantly the TUPAC dataset~\cite{veta2019predicting}, for which an alternative ground truth label exists~\cite{bertram2020pathologist}, and the datasets of the MIDOG 2021~\cite{aubreville2023mitosis} and MIDOG 2022 challenges \cite{aubreville2024domain}. These datasets have fostered extensive research on this pattern recognition task and comparison of numerous deep learning approaches. 

Mitosis is the process of cell division, which is highly regulated in normal cells, resulting in the histological appearance of normal/typical MFs (NMF). However, in malignant tumor cells, mutations can disrupt this regulated process. Resultant failure of equal division of the genetic material (chromosomes) between the two daughter cells can be detected in histological images as atypical MFs (AMFs, Fig.~\ref{3644-fig1}). Aberrant chromosome constitutions (aneuploidy) of neoplastic cells can promote tumor progression~\cite{gisselsson2008classification}. There is increasing evidence that the number of AMFs (AMF count) and the ratio of AMFs among all MFs is of high prognostic relevance in human breast cancer \cite{lashen2022characteristics,ohashi2018prognostic} and other tumor types \cite{bertram2023atypical,matsuda2016mitotic}. The inter-rater agreement of classifying MFs into normal and atypical subtypes was reported as low due to the complex and highly variable morphologies~\cite{bertram2023atypical,aubreville2023deep}. 
Deep learning models have the potential to improve reproducibility and lower time investment required for this task. 

To date there are only two studies that have investigated deep learning models for AMFs \cite{aubreville2023deep,fick2024improving}, 
both of which have shown that this task seems to be particularly difficult based on the overlapping morphologies of normal MFs and AMFs as well as the low frequency of AMFs (class imbalance). Open access datasets are needed to facilitate development of optimized deep learning approaches. 
Following common practice, a reasonable algorithmic workflow seems to be a two step process with first an object detection of all MFs within large images and a second patch classification into normal and atypical subtypes. Whereas this first step has been extensively addressed in previous research (see above), this article focuses on the patch classification task. 

The aim of this dataset publication is to make the first 
atypical mitotic figure dataset for human breast cancer (AMi-Br) publicly available and provide baseline performance for deep learning classification models. 

\section{Materials and methods}

\subsection{Dataset creation}
For this dataset we used all MFs labels from the two largest and most diverse challenge datasets in human breast cancer, namely the TUPAC challenge \cite{veta2019predicting} using the improved alternative version of labels \cite{bertram2020pathologist} and the MIDOG 2021 challenge \cite{aubreville2023mitosis}. These two datasets comprise histological sections from three pathology centers scanned with 6 different whole slide image scanners.
A preliminary version of this dataset has been created for a previous study \cite{aubreville2023deep}, whereas a third annotator was added to derive a consensus vote on the labels. 

\begin{figure}
    \centering
    \includegraphics[width=1\linewidth]{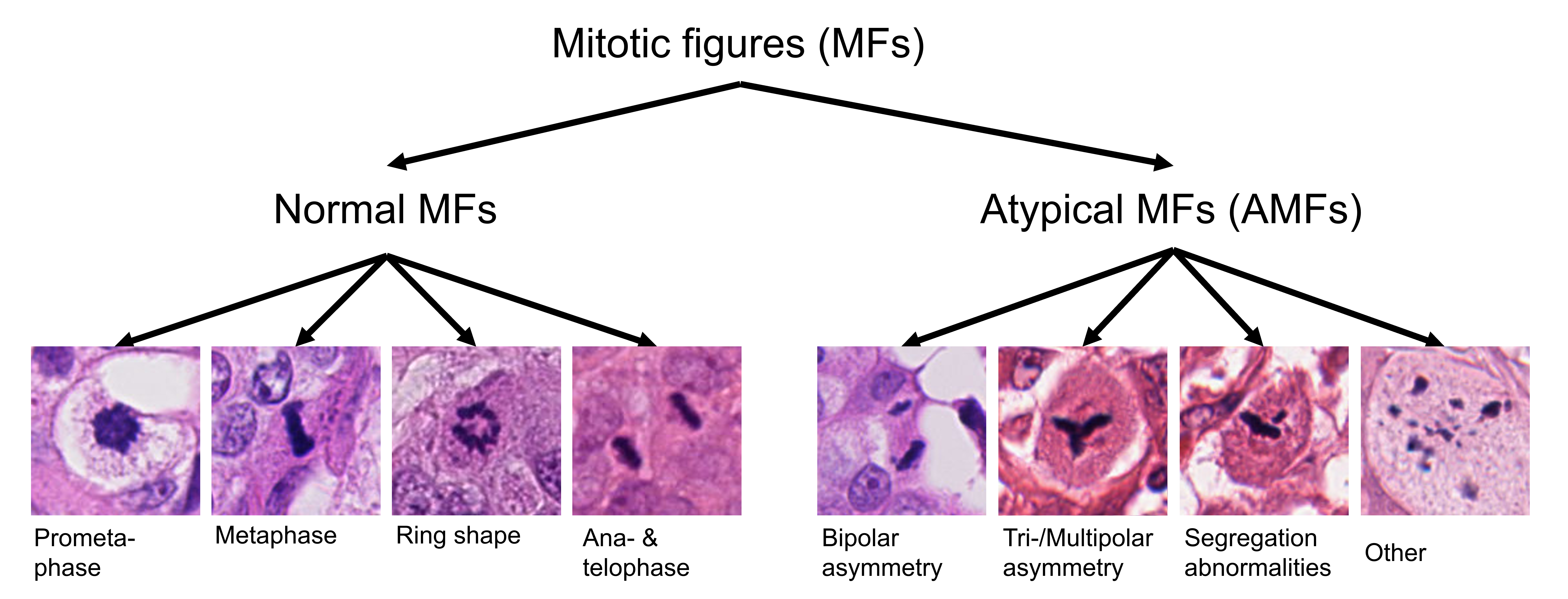}
    \caption{Examples of the mitotic figure morphologies included in the AMi-Br dataset.}
    \label{3644-fig1}
\end{figure}

All three annotators (CAB, VW and TAD) received image patches (128 $\times$ 128 pixels) centered around the original MF annotations and were asked to independently (blinded to the other annotators) classify them into normal and atypical morphologies (Fig.~\ref{3644-fig1}). Normal MFs were defined by the characteristic morphologies of the different phases (1. prometaphase, 2. metaphase, 3. ring-shape metaphase, 4. ana- and telophase), as previously described \cite{donovan2021mitotic}. AMFs included the following categories: 1. polar asymmetry  (1A. bipolar asymmetry, 1B. tri- and multipolar asymmetry), 2. chromosome segregation abnormalities (including lagging chromosomes and bridging chromosomes) and 3. other atypical morphologies (such as dispersed chromosome fragments), based on previous definitions~\cite{lashen2022characteristics,ohashi2018prognostic,bertram2023atypical}. If segregation abnormalities occurred concurrently  with polar asymmetry or other AMFs, the MF was assigned to the latter label class. For MF patches that lacked clear characteristics of the above mentioned label classes, annotators had to select the class with the highest resemblance. Information relating to the specific MF and AMF subtypes is available in the dataset for two pathologists, however is not  considered for the technical validation as it is difficult to find a majority vote for 8 label classes. 

\subsection{Technical validation}
To validate our dataset and assess the label consistency, we performed baseline classification experiments using a standard pipeline with DenseNet-121 and EfficientNet V2 S (both pre-trained on ImageNet) as backbones and using several strategies to combat the class imbalance. We first used weighted cross-entropy, using the class occurrences as weights. Additional strategies included testing random class prevalence-weighted sampling with replacement and using the focal loss. We trained all models for 20 epochs using the Adam optimizer with a learning rate of $10^{-3}$ and retrospectively selected the best model on the balanced accuracy on the validation set.

As the problem is imbalanced, we report balanced accuracy as well as area under the ROC curve (ROC AUC). We performed a five-fold Monte Carlo cross-validation for each of the conditions and report mean and standard deviation values for each metric. 
We employed two distinct strategies for data splitting. First, to evaluate the quality of the labels, we conducted the split at the patch level. While we acknowledge that this approach may yield overoptimistic results, we additionally performed a split at the patient level to establish a baseline for algorithmic comparisons.

\subsection{Usage notes}
We make the full dataset available in our github repository\footnote{\url{https://github.com/DeepMicroscopy/AMi-Br/}}. The dataset is stored as CSV file, encompassing the dataset, filename, coordinates, and individual expert labels as well as the joint (majority vote) atypical label as the columns. The repository further contains the notebooks of our experiments. 
 
\section{Results}

\subsection{Dataset description}
The dataset consists of, in total, 3,720 MFs, comprising 1,999 MFs from the TUPAC16 alternative label set and 1,721 MFs from the MIDOG21 dataset. 

\paragraph{Atypical vs. typical}
As the result from the majority vote between the three experts, 832 MFs were found to be 
atypical, while the remaining 2,888 were considered normal MFs. 
All experts agreed on 2,908 objects, 412 of which were atypical MFs. The agreement of individual experts can be found in Table \ref{3644tab:experts}. 

\paragraph{Morphology}
Overall, we found a moderate agreement ($\kappa=0.60$) between experts 1 and 2 regarding the morphological subclassification of MFs (Tables \ref{3644tab:experts} and \ref{3644tab:confusion}). In the mitotic phase classification of normal MFs, we found the main discrepancy between adjacent classes in the mitotic cycle (Table \ref{3644tab:confusion}).

\begin{table}[]
    \centering
    \caption{Assessment of experts' agreement (overall agreement and Cohen's $\kappa$) on the atypical classification and the morphology classification (4 normal MF (NMF) phases and 4 AMF subtypes).}
\begin{tabular}{p{2cm}p{2cm}cc}
\hline
 evaluation & experts & agreement & Cohen's $\kappa$\\
\hline
\multirow{3}{*}{AMF/NMF} & 1 vs. 2 &  0.85 & 0.53 \\
 & 1 vs. 3 &  0.84 & 0.54 \\
 & 2 vs. 3 &  0.87 & 0.66 \\
\hline
morphology & 1 vs. 2 &  0.71 & 0.60 \\
\hline
\end{tabular}
    \label{3644tab:experts}
\end{table}

\begin{table}
\caption{Confusion matrix between expert 1 and expert 2 in the assessment of MF morphology.}
\resizebox{\linewidth}{!}{
\begin{tabular}{lrrrrrrrr}
\hline
 & \multicolumn{4}{c}{AMF} & \multicolumn{4}{c}{NMF}\\
 & bipolar asym. & multipolar & other & segregation & anaphase-telophase & metaphase & prometaphase & ring shape \\
 
 &  &  &  &  &  &  &  &  \\
\hline
AMF bipolar asym. & 35 & 1 & 5 & 9 & 2 & 6 & 1 & 0 \\
AMF multipolar & 2 & 28 & 23 & 11 & 5 & 1 & 1 & 0 \\
AMF other & 9 & 6 & 121 & 46 & 2 & 52 & 15 & 4 \\
AMF segregation & 2 & 0 & 12 & 120 & 3 & 5 & 11 & 1 \\
NMF anaphase-telophase & 32 & 0 & 7 & 32 & 166 & 14 & 2 & 0 \\
NMF metaphase & 7 & 2 & 12 & 121 & 10 & 1175 & 22 & 0 \\
NMF prometaphase & 17 & 2 & 53 & 131 & 19 & 244 & 949 & 112 \\
NMF ring shape & 1 & 0 & 5 & 9 & 1 & 2 & 0 & 34 \\
\hline
\end{tabular}}
\label{3644tab:confusion}
\end{table}

\subsection{Technical validation}

\begin{table}[]
\setlength{\tabcolsep}{1.5em} 
    \centering
        \caption{Results of the classification experiment. We report mean $\pm$ standard deviation of five-fold Monte Carlo cross-validation for each metric on the resp. test set.}    
\resizebox{\textwidth}{!}{ %
\begin{tabular}{llcccc}
    \hline
\multirow{2}{*}{model} & \multirow{2}{*}{strategy} & \multicolumn{2}{c}{patch-level split} & \multicolumn{2}{c}{patient-level split} \\
& & bal. acc. & ROC AUC & bal. acc. & ROC AUC \\
    \hline
\multirow{3}{*}{DenseNet-121} & weighted cross-entropy &0.801 $\pm$ 0.019 & 0.879 $\pm$ 0.010 & 0.617 $\pm$ 0.031 & 0.633 $\pm$ 0.037\\
& focal loss &0.791 $\pm$ 0.018 & 0.899 $\pm$ 0.012 & 0.551 $\pm$ 0.024 & 0.587 $\pm$ 0.064\\
& weighted sampling &0.795 $\pm$ 0.030 & 0.877 $\pm$ 0.015 & 0.651 $\pm$ 0.015 & 0.656 $\pm$ 0.022\\
\hline
\multirow{3}{*}{EfficientNet V2-S} & weighted cross-entropy &0.806 $\pm$ 0.024 & 0.887 $\pm$ 0.011 & 0.675 $\pm$ 0.020 & 0.656 $\pm$ 0.025\\
 & focal loss &0.785 $\pm$ 0.015 & 0.880 $\pm$ 0.013 & 0.671 $\pm$ 0.020 & 0.674 $\pm$ 0.012\\
 & weighted sampling &0.789 $\pm$ 0.016 & 0.876 $\pm$ 0.011 & 0.713 $\pm$ 0.016 & 0.698 $\pm$ 0.026\\
 
    \hline
    \end{tabular}}

    \label{3644tab:results}
\end{table}

Using a patch-level split, we found an average balanced accuracy and ROC AUC of up to 0.806 and up to 0.899, respectively, depending on the strategy used to combat the class imbalance ( Table~\ref{3644tab:results}).  We found a significant drop in performance when patient-level splitting was performed, hinting towards a possible patient-specific prior that the model overfitted to.
We only found a minor impact of model architecture and imbalance-countering strategy, with a slight benefit for the weighted cross-entropy loss and weighted sampling.

\section{Discussion}

As shown by the statistics regarding expert agreement as well as the baseline experiments, AMF classification is a challenging task. The difficulty of this endeavor can be explained by 1) the high class imbalance and 2) the high morphological overlap between AMFs and normal MFs. Since erroneous cell division may occur at any phase of the cell cycle, there may be a higher resemblance between AMFs and normal MFs than within these label classes. Furthermore, some of the atypical morphologies are subtle, such as lagging chromosomes (i.e. small chromosome fragment left behind in the cytoplasm near the metaphase plate instead of being pulled apart). 
Algorithmic performance may be improved by annotating even larger datasets, more reproducible AMF definitions, or in regularization strategies like creating dedicated augmentation schemes for this problem. 
Future research should create an AMF dataset for additional tumor types that would allow a wider research application \cite{aubreville2024domain}. While there are few studies that have evaluated AMFs in tumor types other than human breast cancer \cite{bertram2023atypical,matsuda2016mitotic,travaglino2021prognostic}, this test needs to be evaluated more extensively to gain a better understanding of its prognostic role across tumor types. Deep learning models would allow a reproducible evaluation and analysis of entire tumor sections, thereby generating an understanding of the AMF's intratumoral distribution.  

\begin{acknowledgement}
CAB, VW, and CK acknowledge the support from the Austrian Research Fund (FWF, project number: I 6555). SB, TC, RK, and MA acknowledge support by the Deutsche Forschungsgemeinschaft (DFG, German Research Foundation, project number: 520330054).
\end{acknowledgement}

\printbibliography

@article{bertram2023atypical,
  title={Atypical mitotic figures are prognostically meaningful for canine cutaneous mast cell tumors},
  author={Bertram, Christof A and Bartel, Alexander and Donovan, Taryn A and Kiupel, Matti},
  journal={Veterinary Sciences},
  volume={11},
  number={1},
  pages={5},
  year={2023},
  publisher={MDPI}
}

@inproceedings{aubreville2023deep,
  title={Deep learning-based Subtyping of Atypical and Normal Mitoses using a Hierarchical Anchor-Free Object Detector},
  author={Aubreville, Marc and Ganz, Jonathan and Ammeling, Jonas and Donovan, Taryn A and Fick, Rutger HJ and Breininger, Katharina and Bertram, Christof A},
  booktitle={BVM Workshop},
  pages={189--195},
  year={2023},
  organization={Springer}
}

@article{aubreville2024domain,
  title={Domain generalization across tumor types, laboratories, and species—Insights from the 2022 edition of the Mitosis Domain Generalization Challenge},
  author={Aubreville, Marc and Stathonikos, Nikolas and Donovan, Taryn A and Klopfleisch, Robert and Ammeling, Jonas and Ganz, Jonathan and Wilm, Frauke and Veta, Mitko and Jabari, Samir and Eckstein, Markus and others},
  journal={Medical Image Analysis},
  volume={94},
  pages={103155},
  year={2024},
  publisher={Elsevier}
}

@article{aubreville2023mitosis,
  title={Mitosis domain generalization in histopathology images—the MIDOG challenge},
  author={Aubreville, Marc and Stathonikos, Nikolas and Bertram, Christof A and Klopfleisch, Robert and Ter Hoeve, Natalie and Ciompi, Francesco and Wilm, Frauke and Marzahl, Christian and Donovan, Taryn A and Maier, Andreas and others},
  journal={Medical Image Analysis},
  volume={84},
  pages={102699},
  year={2023},
  publisher={Elsevier}
}

@inproceedings{bertram2020pathologist,
  title={Are pathologist-defined labels reproducible? Comparison of the TUPAC16 mitotic figure dataset with an alternative set of labels},
  author={Bertram, Christof A and Veta, Mitko and Marzahl, Christian and Stathonikos, Nikolas and Maier, Andreas and Klopfleisch, Robert and Aubreville, Marc},
  booktitle={Interpretable and Annotation-Efficient Learning for Medical Image Computing},
  pages={204--213},
  year={2020},
  organization={Springer}
}

@article{veta2019predicting,
  title={Predicting breast tumor proliferation from whole-slide images: the TUPAC16 challenge},
  author={Veta, Mitko and Heng, Yujing J and Stathonikos, Nikolas and Bejnordi, Babak Ehteshami and Beca, Francisco and Wollmann, Thomas and Rohr, Karl and Shah, Manan A and Wang, Dayong and Rousson, Mikael and others},
  journal={Medical image analysis},
  volume={54},
  pages={111--121},
  year={2019},
  publisher={Elsevier}
}

@inproceedings{fick2024improving,
  title={Improving CNN-Based Mitosis Detection through Rescanning Annotated Glass Slides and Atypical Mitosis Subtyping},
  author={Fick, Rutger RH and Bertram, Christof and Aubreville, Marc},
  booktitle={Medical Imaging with Deep Learning},
  year={2024}
}

@article{bertram2024mitotic,
  title={Mitotic activity: A systematic literature review of the assessment methodology and prognostic value in canine tumors},
  author={Bertram, Christof A and Donovan, Taryn A and Bartel, Alexander},
  journal={Veterinary pathology},
  pages={03009858241239565},
  year={2024},
  publisher={SAGE Publications Sage CA: Los Angeles, CA}
}

@article{van2022grading,
  title={Grading of invasive breast carcinoma: the way forward},
  author={Van Dooijeweert, C and Van Diest, PJ and Ellis, IO},
  journal={Virchows Archiv},
  volume={480},
  number={1},
  pages={33--43},
  year={2022},
  publisher={Springer}
}

@article{gisselsson2008classification,
  title={Classification of chromosome segregation errors in cancer},
  author={Gisselsson, David},
  journal={Chromosoma},
  volume={117},
  number={6},
  pages={511--519},
  year={2008},
  publisher={Springer}
}

@article{ohashi2018prognostic,
  title={Prognostic utility of atypical mitoses in patients with breast cancer: A comparative study with Ki67 and phosphohistone H3},
  author={Ohashi, Ryuji and Namimatsu, Shigeki and Sakatani, Takashi and Naito, Zenya and Takei, Hiroyuki and Shimizu, Akira},
  journal={Journal of surgical oncology},
  volume={118},
  number={3},
  pages={557--567},
  year={2018},
  publisher={Wiley Online Library}
}

@article{lashen2022characteristics,
  title={The characteristics and clinical significance of atypical mitosis in breast cancer},
  author={Lashen, Ayat and Toss, Michael S and Alsaleem, Mansour and Green, Andrew R and Mongan, Nigel P and Rakha, Emad},
  journal={Modern Pathology},
  volume={35},
  number={10},
  pages={1341--1348},
  year={2022},
  publisher={Elsevier}
}

@article{matsuda2016mitotic,
  title={Mitotic index and multipolar mitosis in routine histologic sections as prognostic markers of pancreatic cancers: a clinicopathological study},
  author={Matsuda, Yoko and Yoshimura, Hisashi and Ishiwata, Toshiyuki and Sumiyoshi, Hiroki and Matsushita, Akira and Nakamura, Yoshiharu and Aida, Junko and Uchida, Eiji and Takubo, Kaiyo and Arai, Tomio},
  journal={Pancreatology},
  volume={16},
  number={1},
  pages={127--132},
  year={2016},
  publisher={Elsevier}
}

@article{donovan2021mitotic,
  title={Mitotic figures—normal, atypical, and imposters: A guide to identification},
  author={Donovan, Taryn A and Moore, Frances M and Bertram, Christof A and Luong, Richard and Bolfa, Pompei and Klopfleisch, Robert and Tvedten, Harold and Salas, Elisa N and Whitley, Derick B and Aubreville, Marc and others},
  journal={Veterinary pathology},
  volume={58},
  number={2},
  pages={243--257},
  year={2021},
  publisher={SAGE Publications Sage CA: Los Angeles, CA}
}

@article{travaglino2021prognostic,
  title={Prognostic significance of atypical mitotic figures in smooth muscle tumors of uncertain malignant potential (STUMP) of the uterus and uterine adnexa},
  author={Travaglino, Antonio and Raffone, Antonio and Santoro, Angela and Gencarelli, Annarita and Angelico, Giuseppe and Spadola, Saveria and Marzullo, Liberato and Zullo, Fulvio and Insabato, Luigi and Zannoni, Gian Franco},
  journal={Apmis},
  volume={129},
  number={4},
  pages={165--169},
  year={2021},
  publisher={Wiley Online Library}
}

\end{document}